\documentclass[sigconf]{acmart}

\copyrightyear{2025}
\acmYear{2025}
\setcopyright{cc}
\setcctype{by-nc}
\acmConference[SIGSPATIAL '25]{The 33rd ACM International Conference on Advances in Geographic Information Systems}{November 3--6, 2025}{Minneapolis, MN, USA}
 \acmBooktitle{The 33rd ACM International Conference on Advances in Geographic Information Systems (SIGSPATIAL '25), November 3--6, 2025, Minneapolis, MN, USA}
 \acmDOI{10.1145/3748636.3763220}
 \acmISBN{979-8-4007-2086-4/2025/11}

\begin{document}

\title{Enhancing Contrastive Learning for Geolocalization by Discovering Hard Negatives on Semivariograms}

\author{Boyi Chen}
\email{boyi.chen@tum.de}
\affiliation{%
  \institution{Professorship Big Geospatial Data Management, Technical University of Munich \country{Germany}}
}

\author{Zhangyu Wang}
\authornote{Corresponding author.}
\email{zhangyu.wang@maine.edu}
\affiliation{%
  \institution{University of Maine \country{United States}}
}

\author{Fabian Deuser}
\email{fabian.deuser@unibw.de}
\affiliation{%
  \institution{University of the Bundeswehr Munich \country{Germany}}
}

\author{Johann Maximilian Zollner}
\email{maximilian.zollner@tum.de}
\affiliation{%
  \institution{Professorship Big Geospatial Data Management, Technical University of Munich \country{Germany}}
}

\author{Martin Werner}
\email{martin.werner@tum.de}
\affiliation{%
  \institution{Professorship Big Geospatial Data Management, Technical University of Munich \country{Germany}}
}



\renewcommand{\shortauthors}{Chen et al.}

\begin{abstract}



Accurate and robust image-based geo-localization at a global scale is challenging due to diverse environments, visually ambiguous scenes, and the lack of distinctive landmarks in many regions. While contrastive learning methods show promising performance by aligning features between street-view images and corresponding locations, they neglect the underlying spatial dependency in the geographic space. As a result, they fail to address the issue of false negatives - image pairs that are both visually and geographically similar but labeled as negatives, and struggle to effectively distinguish hard negatives, which are visually similar but geographically distant. To address this issue, we propose a novel spatially regularized contrastive learning strategy that integrates a semivariogram, which is a geostatistical tool for modeling how spatial correlation changes with distance. We fit the semivariogram by relating the distance of images in feature space to their geographical distance, capturing the expected visual content in a spatial correlation. With the fitted semivariogram, we define the expected visual dissimilarity at a given spatial distance as reference to identify hard negatives and false negatives. We integrate this strategy into GeoCLIP and evaluate it on the OSV5M dataset, demonstrating that explicitly modeling spatial priors improves image-based geo-localization performance, particularly at finer granularity.
\end{abstract}



\maketitle
\section{Introduction}
\label{sec:intro}

\begin{figure*}
    \centering
    \includegraphics[width=0.90\linewidth]{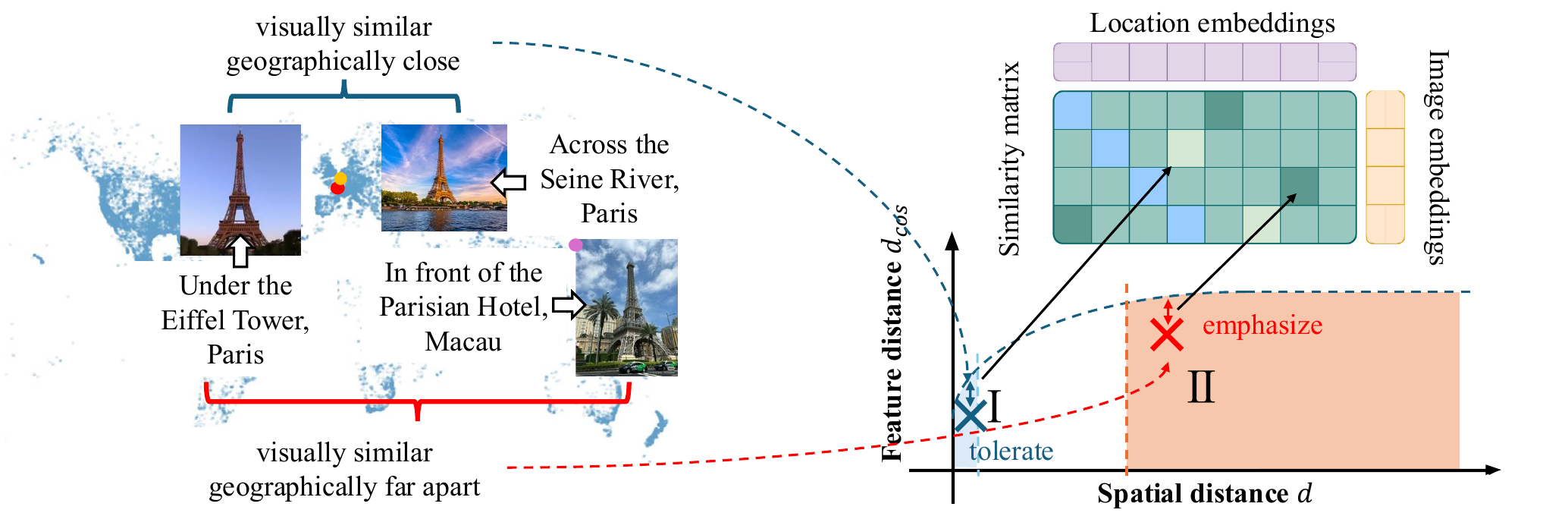}
    \caption{Spatial-aware reweighting strategy. Based on the semivariogram, we (1) emphasize visually similar but geographically distant pairs (hard negatives), and (2) tolerate visually similar and geographically close pairs (false negatives).}
    \vspace{-0.2cm}
    \label{fig:enter-label}
\end{figure*}

Image-based geo-localization aims to determine the geographic location of a query image. This is particularly useful in scenarios where Global Positioning System (GPS) signals are unavailable or unreliable, such as dense urban areas with tall buildings or mountainous regions. It has a wide range of applications in disaster response and augmented reality \cite{li2025cross, mithun2023cross}. However, image-based geo-localization at the global scale is a critical challenge in computer vision and GeoAI due to the diversity of visual appearances across the world and the lack of distinctive landmarks in many regions. 

Recently, contrastive learning has significantly advanced the field of image-based geo-localization by enabling models to capture rich semantic information that is crucial for recognizing locational cues in images. Contrastive Language-Image Pre-Training (CLIP) \cite{clip} has become a powerful learning paradigm in geo-localization tasks, especially at the global scale, where representations can be learned by leveraging the supervision of large-scale geo-tagged imagery. By encouraging semantically similar pairs to be close in the embedding space and pushing dissimilar pairs apart, contrastive learning effectively aligns visual inputs with their corresponding labels, whether textual descriptions or geographic coordinates.

Despite these advances, current approaches mainly treat geographic coordinates as pure numerical inputs and ignore their interactions with the observed images. They do not consider the fact that images geographically near to each other are often visually similar due to spatial autocorrelation, a phenomenon explained by Tobler's First Law of Geography \cite{tobler1970computer}, which states that "everything is related to everything else, but near things are more related than distant things". This spatial autocorrelation remains under-utilized in current geo-localization models. Furthermore, although contrastive learning architectures such as CLIP benefit from hard negatives by promoting more discriminative representations, the visually similar samples may not be hard negatives but potentially false negatives. Hence, properly identifying and handling these negatives is a key to improving the performance of contrastive learning models.


Unlike general vision tasks, geo-localization involves spatial data, where similarity depends on distance-dependent patterns. Thus, we propose to integrate the geographic information into the contrastive learning loss with a semivariogram, to make the model aware of the geographic reality. The semivariogram is a geostatistical tool that models the expected dissimilarity between samples as a function of their geographic distances. We model the expected relation in street view images across different regions, which allows us to distinguish between hard negatives and false negatives. Consequently, we penalize hard negatives during training to reduce the impact of false negatives on the performance of the model.

\textbf{Contributions}. In summary, our contributions are: (1) We introduce a semivariogram-based approach to model spatial autocorrelation in image similarity, demonstrating that geographic distance and visual similarity are systematically related at a global scale. (2) We design a semivariogram-guided reweighting strategy within contrastive learning to explicitly identify and handle hard negatives and false negatives, improving the model’s spatial awareness.
\section{Related Work}
\label{sec:related_work}

Image-based geo-localization methods can be broadly categorized into classification-based and retrieval-based approaches. Classification-based methods, such as PlaNet \cite{weyand2016planet}, divide the Earth's surface into a set of cells and treat the localization task as an image classification problem, the model predicts which geographic cell the input image most likely belongs to. Retrieval-based methods like Im2GPS \cite{vo2017revisiting} match the query image with a reference database to query the most similar one as prediction. StreetCLIP \cite{haas2023learning} adapts CLIP to geolocalization by matching images to a batch of synthetically generated location descriptions. During inference, the model predicts the location whose description yields the highest similarity to the image embedding, enabling effective zero-shot localization. 
Sample4Geo \cite{deuser2023sample4geo} using a symmetric InfoNCE contrastive learning loss with a novel hard negative sampling, selecting samples from both geographic proximity and visual similarity, leading to state-of-the-art performance cross-view geolocalization. Generative methods \cite{dufour2025around} propose a probabilistic formulation using diffusion and flow matching, with significant improvement at large scales.

\section{Method}
\label{sec:method}

The proposed method introduces a negative mining strategy based on semivariogram information. We test its effectiveness with a contrastive learning architecture GeoCLIP \cite{vivanco2023geoclip}. Instead of treating all negative samples equally, we model spatial autocorrelation across the geographic space by fitting a theoretical semivariogram between the feature dissimilarities of image embeddings and their geographic distances.  
The fitted semivariogram defines a relationship between the spatial distance from an anchor image to its negative samples and their expected visual dissimilarity. Negative samples that significantly deviate from this relationship are considered to be potential hard negatives/false negatives. During training, we put more emphasis (i.e., higher weights) on hard negatives and less emphasis on false negatives. 

\subsection{GeoCLIP}
\label{subset:geoclip}

GeoCLIP consists of two main components: an image encoder $\mathcal{V}(\cdot)$ and a location encoder $\mathscr{L}(\cdot)$. The image encoder is based on a frozen pre-trained CLIP image encoder ViT-L/14 \cite{dosovitskiy2020image}, following a two-layer trainable MLP to fine-tune the extracted features. The location encoder first applies the Equal Earth Projection (EEP)~\cite{vsavrivc2019equal} to reduce the distortion caused by unequal division of longitude and latitude. In order to capture rich details at different scales, GeoCLIP constructs a hierarchical representation using Random Fourier Features \cite{tancik2020fourier} with $M$ different frequencies $\sigma$ in a range from $2^0$ to $2^8$. Then the encoded hierarchical features are passed through a trainable MLP to learn the suitable embeddings. Let the Random Fourier Feature transformation be denoted as $\gamma$ and the location MLP as $f$. Given a geo-tagged image $(I_i, G_i)$ where $I_i$ is an image and $G_i$ is its coordinate, the image encoder $\mathcal{V}(\cdot)$ and the location encoder $\mathscr{L}(\cdot)$ for the feature extraction can then be written as:

\begin{equation}
    V_i = \mathcal{V}(I_i)     
\end{equation}

\begin{equation}
    L_i = \mathscr{L}(G_i) = \sum_{k=1}^M f_k(\gamma(EEP(G_i), \sigma_k)
\end{equation}

During each step of training, we draw a batch of size $B$, $\mathcal{B} = \{(I_i, G_i)\}_{i=1}^{B}$, from the training dataset $\mathcal{D}_{train} = \{(I_n, G_n)\}_{n=1}^N$. For the $i^{th}$ sample $(V_{i}, L_i)$ in batch $\mathcal{B}$, the negative set $\mathcal{N}_i$ is defined as $(\mathcal{B} - \{(V_{i}, L_i)\}) \bigcap \mathcal{Q}$, where $\mathcal{Q}$ is a dynamic queue of geo-tagged images constructed from $\mathcal{D}_{train}$ as is described in \cite{vivanco2023geoclip}. For each batch, the training objective is to minimize the InfoNCE loss \cite{oord2018representation}:



\begin{equation}
\label{eq:geocliploss}
    \sum_{i=1}^{B}\mathcal{L}_i = \sum_{i=1}^{B}-\log\frac{\exp(V_{i}\cdot L_{i}/\tau)}{\exp(V_{i}\cdot L_{i}/\tau) + \sum_{{L}^-_{j} \in \mathcal{N}_i} \exp(V_{i} \cdot {L}^-_{j}/\tau)}
\end{equation}

In practice, we use the average loss computed on $P=2$ augmentations of each image to mitigate overfitting, as is a common trick used in contrastive learning \cite{chen2020simple}.


\subsection{Semivariogram}
\label{subset:semi}

Given a finite set of locations $G$ with associated univariate observations $z$, the classical semivariogram \cite{matheron1963principles} models the spatial dependency by quantifying how the dissimilarity between observations increases as their spatial distance increases. The empirical semivariogram at distance $h$ over a small tolerance $\varepsilon$ is defined as:
\begin{equation}\label{eq:classic-semivariogram}
    \hat{\gamma}(h\pm{\mathcal{\varepsilon}}) := \frac{1}{2|N(h\pm{\varepsilon})|}\sum_{(G_{i},G_{j}) \in N(h \pm{\varepsilon})}|z_{i} - z_{j}|^2
\end{equation}

where $N(h\pm{\varepsilon}) := \lbrace (G_{i}, G_{j})| h-\varepsilon \leq d(i, j) \leq h + \varepsilon \rbrace $ denotes the set of point pairs whose geographic distance $d$ is within the threshold $h \pm{ \varepsilon}$ (in our case, $d$ is the great-circle distance from $G_i$ to $G_j$). MC-GTA \cite{wang2024mc} generalized this to multivariate observations (e.g., image embeddings) by replacing the $|z_{i} - z_{j}|^2$ term with Wasserstein-2 distances. This generalization inspires us to adopt semivariograms in our geo-localization setting. We treat the image embeddings $V_i$ as multivariate observations and measure their dissimilarity using cosine distances. Formally, given a pair of image embeddings $(V_i, V_j)$ located at $(G_i, G_j)$, we compute:
\begin{equation}
    d_{\cos}(i, j) = 1 - \frac{V_i \cdot V_j}{||V_i||||V_j||}
\end{equation}
where $\cdot$ denotes the dot product between the two feature vectors to measure the similarity. Replacing the univariate difference $|z_{i} - z_{j}|^2$ in Equation \ref{eq:classic-semivariogram} with this cosine distance, we estimate the \textbf{empirical embedding semivariogram} $\hat{\gamma}_e$ that captures the expected image embedding dissimilarity as a function of geographic distance:

\begin{equation}
        \hat{\gamma}_e(h\pm{\mathcal{\varepsilon}}) := \frac{1}{2|N(d\pm{\varepsilon})|}\sum_{(G_{i},G_{j}) \in N(d \pm{\varepsilon})}d_{\cos}(i, j).
\end{equation}



Then we use the spherical semivariogram model to fit an analytical semivariogram function $\gamma$ from the empirical semivariogram $\gamma_e$, which we will use in the next section.

\subsection{Semivariogram Based Reweighting}
\label{subset:reweight}

In the loss computation \eqref{eq:geocliploss}, all negative pairs are treated equally regardless of their spatial context. However, this neglects the impact of false negatives and hard negatives in contrastive learning, specifically for geo-localization. To address these challenges, we propose a reweighting mechanism guided by the fitted semivariogram, allowing us to assign higher importance to hard negatives and reduce the impact of false negatives during training.

Let $d_{\cos}(i, j)$ and $d(i, j)$ denote the cosine distance and haversine distance between the negatives to the true sample. From the fitted semivariogram, we can define the expected visual dissimilarity ${\gamma}(d(i, j))$ at this spatial distance $d(i, j)$, and compare it to the calculated dissimilarity $d_{\cos}(i, j)$, yielding the deviation:

\begin{equation}
    \delta(i,j) = d_{\cos}(i,j) - \gamma(d(i,j))
\end{equation}

This deviation quantifies whether a negative sample is more similar or less similar than expected given its spatial distance. We hypothesize that hard negatives are samples with large spatial distance and $\delta(i,j) < 0$, and that false negatives are samples with small spatial distance and $\delta(i,j) < 0$. Based on this hypothesis, we define the weight $w_{ij}$:
\begin{equation} \label{eq:weighting}
    w_{ij} =
    \begin{cases}
    \exp{(-\delta(i,j)/s_1)}, \ \text{if } \delta(i,j) < 0 \ \text{and} \ d(i,j) > \theta_1 \quad  \text{(hard negative)} \\
    \exp{(\delta(i,j)/s_2)}, \ \text{if } \delta(i,j) < 0 \ \text{and} \ d(i,j) < \theta_2 \quad  \text{(false negative)} \\
    1, \ \text{otherwise}
    \end{cases}
\end{equation}
where $s_1, s_2$ are two hyperparameters used to scale $\delta(i,j)$ for the sake of numerical stability. Finally, we reweight the loss function:

\begin{equation}
\label{eq:semiloss}
    \mathcal{L}_i = -\log\frac{\exp(V_{i}\cdot L_{i}/\tau)}{\exp(V_{i}\cdot L_{i}/\tau) + \sum_{{L}_{j}^- \in \mathcal{N}_i} \exp(w_{ij}(V_{i} \cdot L_{j}^-/\tau))}
\end{equation}

Intuitively, Equation \ref{eq:weighting} emphasizes (with weights > 1) the cases when far apart observations look more similar than they should -- these are the negative samples that are difficult to distinguish visually, i.e. hard negatives; it also tolerates (with weights < 1) the cases when near observations look more similar than they should -- both their distance and image similarity indicate that they are unlikely negative samples.

\section{Experiment}
\label{sec:experiment}

\begin{figure}
    \centering
    \includegraphics[width=0.75\linewidth]{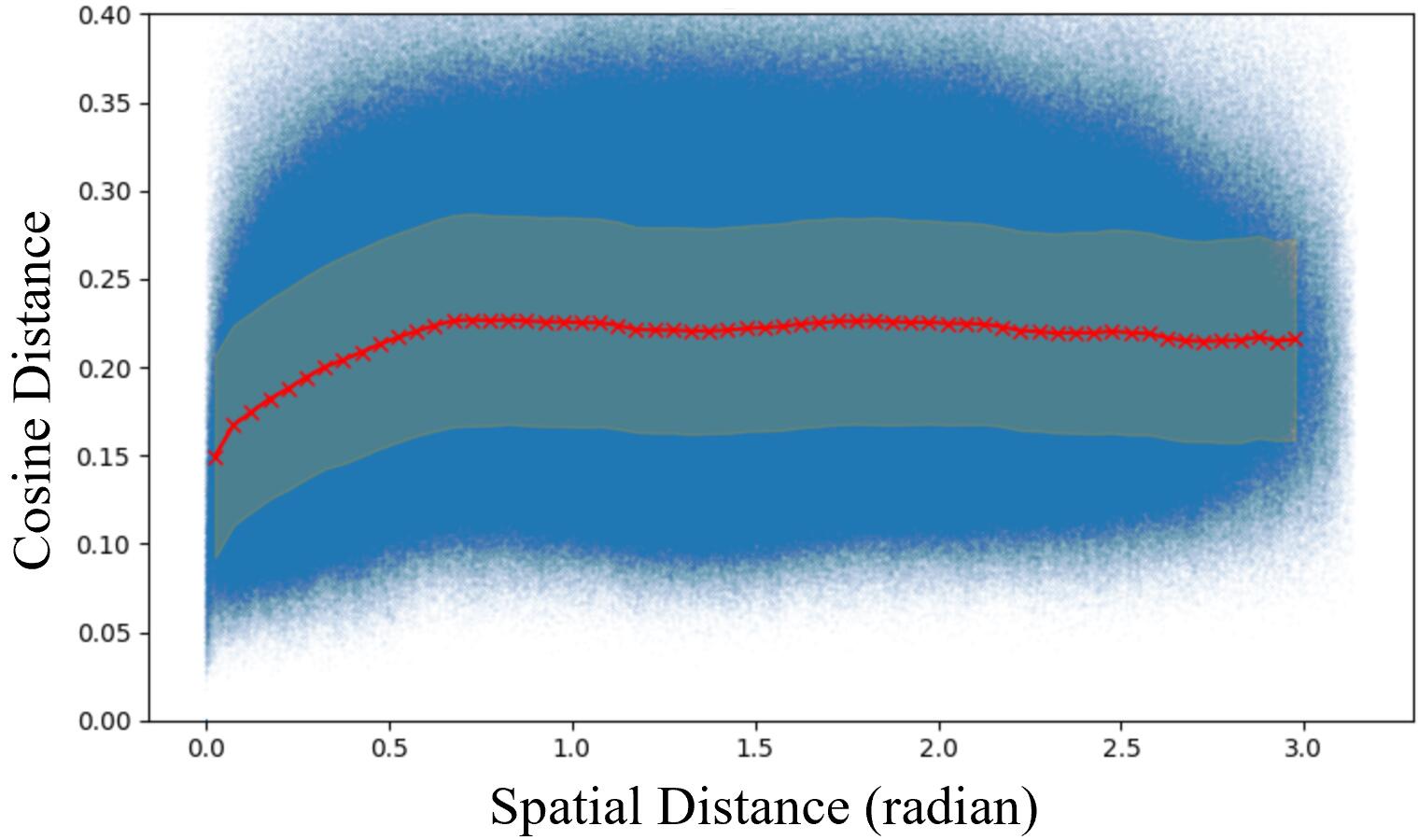}
    \caption{Embedding Semivariogram of OSV5M. Feature distance increases with the spatial distance until a certain range.}
    \vspace{-0.3cm}
    \label{fig:osv5msemi}
\end{figure}

\textbf{Dataset:}
We evaluate our method on the OpenStreetView-5M (OSV5M) dataset \cite{astruc2024openstreetview}, a large-scale benchmark for image-based geo-localization. OSV5M contains around 5 million geo-tagged street-view images sourced from Mapillary. We estimate the empirical semivariogram by computing pairwise cosine distances of image embeddings processed by frozen CLIP image encoder, against their corresponding haversine distances. Shown in Figure \ref{fig:osv5msemi}, illustrates a clear spatial dependency: image dissimilarity increases with increasing geographic distance up to a certain range, beyond which the spatial distance doesn't contribute to the feature distance anymore. 

\textbf{Evaluation:} As a retrieval-based approach, GeoCLIP constructs a GPS gallery for retrieval. The goal is to retrieve the corresponding coordinates from the gallery for a given image. The model computes cosine similarities between the query image embedding and all GPS gallery embeddings, and the coordinate with the highest similarity is selected as the predicted location. Following the common metrics, which evaluate the retrieval accuracy at different levels: $25$km, $200$km, and $750$km. A prediction is considered correct if it falls within the specified threshold radius.

\textbf{Results and Discussion:}
We evaluate the effectiveness of our proposed strategy by integrating it into the GeoCLIP framework. As reported in Table \ref{tab:osv5m_accuracy}, our method consistently outperforms the original GeoCLIP across all evaluation scales. It achieves 52.1\% and 21.5\% accuracy at the region (200km) and city level (25km), improving upon GeoCLIP's 50.1\% and 19.8\%. These gains in fine-grained highlight the benefit of incorporating spatial information into contrastive learning.

Compared to other strong baselines (results from \cite{dufour2025around}), our method achieves the highest region and city level accuracy, while models like RFM perform 4\% better at country level, our method maintains competitive global accuracy while achieving four times higher at city scale. This demonstrates that our semivariogram-guided penalty successfully manage a balance in both coarse and fine level performance.
\vspace{-0.2cm}
\begin{table}[ht]
\centering
\caption{Comparison of geolocalization accuracy on OSV-5M.}
\vspace{-0.2cm}
\begin{tabular}{lccc}
\toprule
Method & \multicolumn{3}{c}{Accuracy ↑ (in \%)} \\
\cmidrule(lr){2-4}
& City(25km) & Region(200km) & Country(750km) \\
\midrule
ISNs \cite{muller2018geolocation}                   & 4.2 & 39.4 & 66.8 \\
Hybrid \cite{astruc2024openstreetview}                & 5.9 & 39.4 & 68.0 \\
SC Retrieval \cite{haas2023learning}   & 19.9 & {45.8} & {73.4} \\
vMF \cite{izbicki2020exploiting}                                & 0.6 & 17.2 & 52.7 \\
{RFM $\mathbb{S}_2$}\cite{dufour2025around} & 5.4 & 44.2 & \textbf{76,2} \\
GeoCLIP \cite{vivanco2023geoclip}  & 19.8   & 50.1   & 71.4  \\
Ours                   & \textbf{21.5}   & \textbf{52.1}  & 72.1  \\
\bottomrule
\end{tabular}
\label{tab:osv5m_accuracy}
\end{table}
\vspace{-0.5cm}

\section{Conclusion}
\label{sec:conclusion}

In this work, we proposed a spatially aware strategy for contrastive learning in geo-localization. By explicitly modeling geographic information using a generalized semivariogram, we provided a perspective to discover hard negatives and false negatives and address two critical challenges in contrastive learning for localization. Integrated into the GeoCLIP framework, our method achieves improvements at fine-grained scales, notably exceeding the original GeoCLIP and other state-of-the-art baselines at the city and region levels, without losing accuracy at the coarse level. These results demonstrate the value of incorporating spatial dependency into representation learning.

Our findings suggest that geographically-aware training can significantly enhace GeoAI models. A promising direction for future research is systematically investigate how geospatial training strategies contribute to the performance and generalization of geospatial artificial intelligence.



\bibliographystyle{ACM-Reference-Format}
\bibliography{refs}

\end{document}